\documentclass[10pt,twocolumn,letterpaper]{article}

\usepackage[pagenumbers]{cvpr} % To force page numbers, e.g. for an arXiv version

\definecolor{cvprblue}{rgb}{0.21,0.49,0.74}
\usepackage[pagebackref,breaklinks,colorlinks,allcolors=cvprblue]{hyperref}
\usepackage{ulem}
\usepackage{bm}

\usepackage{float}
\usepackage{adjustbox}
\usepackage{multirow}
\usepackage{threeparttable}
\usepackage{tablefootnote}
\usepackage{longtable}
\usepackage{pifont}
\usepackage{xspace}
\usepackage{colortbl}
\usepackage{xcolor}
\usepackage{amsmath, amssymb, amsfonts}
\usepackage{mathtools}   % for multline, aligned, cases, 
\usepackage{bm} % 粗体数学符号

\def\paperID{14} % *** Enter the Paper ID here
\def\confName{CVPR}
\def\confYear{2026}

\title{Coffee: \underline{Co}ntrollable Di\underline{ff}usion Fin\underline{e}-tuning}

\author{Ziyao Zeng\textsuperscript{1} \quad
Jingcheng Ni\textsuperscript{2} \quad
Ruyi Liu\textsuperscript{1} \quad
Alex Wong\textsuperscript{1} \quad
\vspace{3mm} \\
\textsuperscript{1}Yale University  \quad
\textsuperscript{2}Brown University \\ \\ 
\tt\small \textsuperscript{1}\{ziyao.zeng,
ruyi.liu, alex.wong\}@yale.edu 
\tt\small \textsuperscript{2}jingcheng\_ni@brown.edu\\
}
\newcommand{\method}{Coffee\xspace}
\begin{document}
\maketitle

%%%%%%%%%%%%%%%%%%%%%%%%%%%%%%%%
%%%%%%%%%%% Abstract %%%%%%%%%%%
%%%%%%%%%%%%%%%%%%%%%%%%%%%%%%%%
\begin{abstract}
Text-to-image diffusion models can generate diverse content with flexible prompts, which makes them well-suited for customization through fine-tuning with a small amount of user-provided data. However, controllable fine-tuning that prevents models from learning undesired concepts present in the fine-tuning data—and from entangling those concepts with user prompts—remains an open challenge. It is crucial for downstream tasks like bias mitigation, preventing malicious adaptation, attribute disentanglement, and generalizable fine-tuning of diffusion policy. We propose \method that allows using language to specify undesired concepts to regularize the adaptation process. The crux of our method lies in keeping the embeddings of the user prompt from aligning with undesired concepts. Crucially, \method requires no additional training and enables flexible modification of undesired concepts by modifying textual descriptions. We evaluate \method by fine-tuning on images associated with user prompts paired with undesired concepts. Experimental results demonstrate that \method can prevent text-to-image models from learning specified undesired concepts during fine-tuning and outperforms existing methods. Code will be released upon acceptance.
\end{abstract}

% Text-to-image diffusion models can generate diverse content with flexible prompts, which makes them well-suited for customization through fine-tuning with a small amount of user-provided data. However, controllable fine-tuning that prevents models from learning undesired concepts present in the fine-tuning data—and from entangling those concepts with user prompts—remains an open challenge. It is crucial for downstream tasks like bias mitigation, preventing malicious adaptation, attribute disentanglement, and generalizable fine-tuning of diffusion policy. We propose Coffee that allows using language to specify undesired concepts to regularize the adaptation process. The crux of our method lies in keeping the embeddings of the user prompt from aligning with undesired concepts. Crucially, Coffee requires no additional training and enables flexible modification of undesired concepts by modifying textual descriptions. We evaluate Coffee by fine-tuning on images associated with user prompts paired with undesired concepts. Experimental results demonstrate that Coffee can prevent text-to-image models from learning specified undesired concepts during fine-tuning and outperforms existing methods. Code will be released upon acceptance.

\begin{figure}[t]
  \centering
  \vspace{0.3cm}
\includegraphics[width=0.5\textwidth]{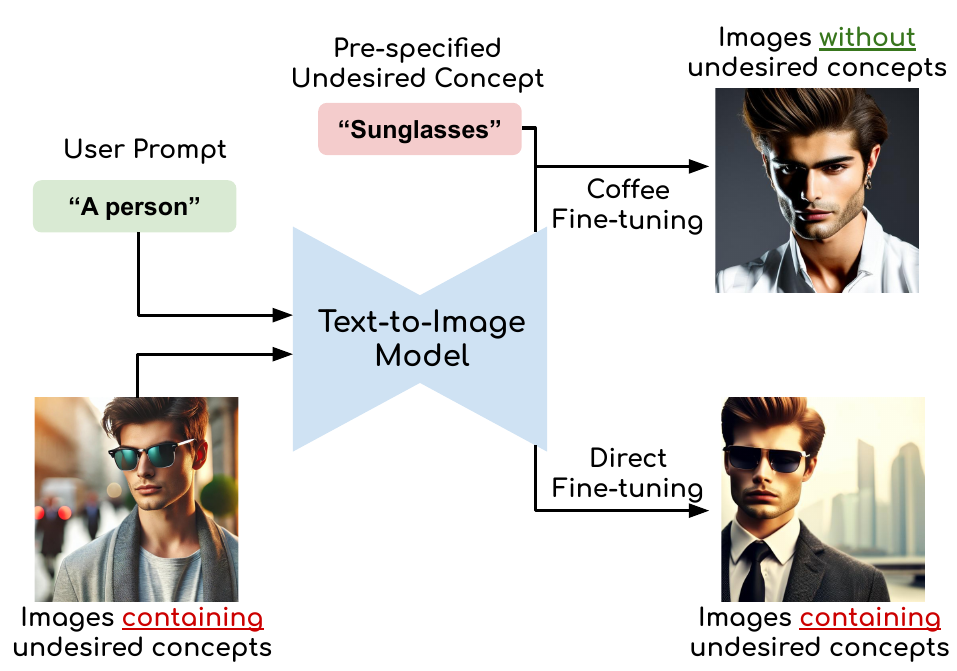}
    \caption{\textbf{Preventing learning undesired concepts during adaptation.} By simply specifying undesired concepts in natural language, \method controls the fine-tuning of the text-to-image diffusion models to prevent the model from learning undesired concepts, and from entangling those concepts with user prompts, when models are fine-tuned with images containing such concepts.
    }
    \label{fig:teaser}
    \vspace{-3mm}
    
\end{figure}

%%%%%%%%%%%%%%%%%%%%%%%%%%%%%%%%%%%%
%%%%%%%%%%% Introduction %%%%%%%%%%%
%%%%%%%%%%%%%%%%%%%%%%%%%%%%%%%%%%%%
\section{Introduction}
\label{sec:intro}
\begin{figure}
    \centering
    \includegraphics[width=\linewidth]{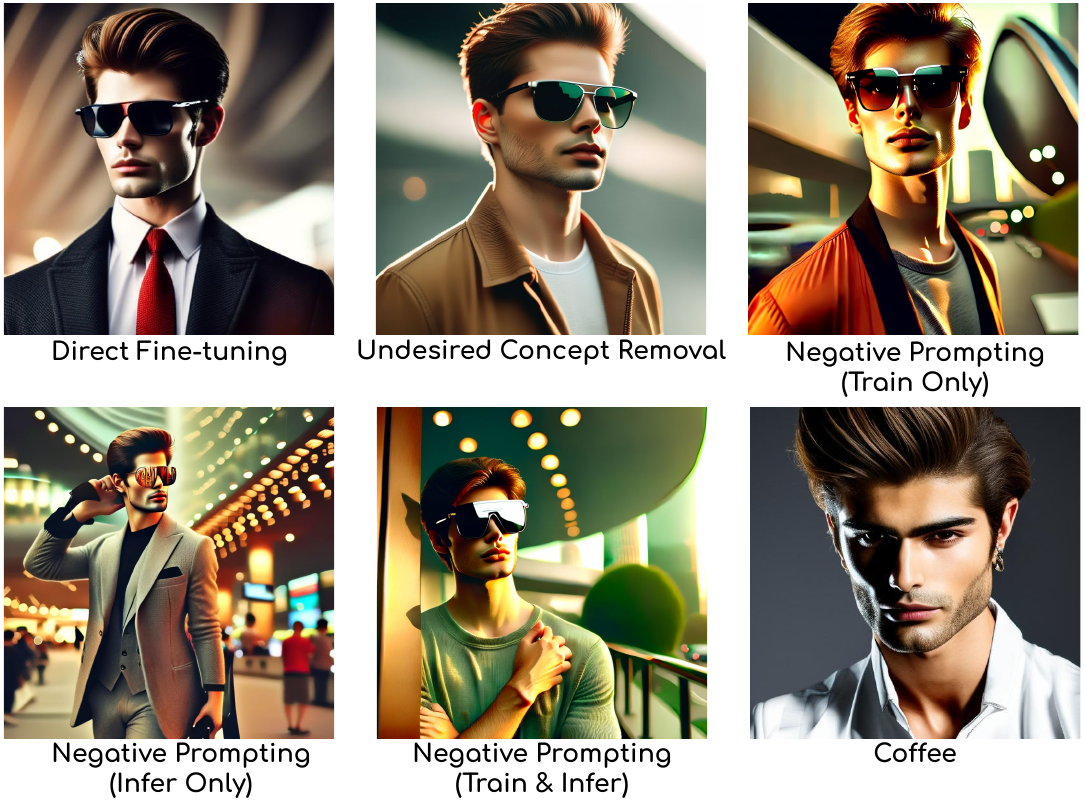}
    \caption{\textbf{Prompt steering alone cannot prevent the model from learning undesired concepts (\textit{``Sunglasses''}).} Negative Prompting is implemented through classifier-free guidance \cite{ho2022classifier}. Undesired Concept Removal is to remove undesired concepts from the input prompt in inference using the format ``\texttt{<user prompt>} without \texttt{<undesired concept>}''. Details can be found in Table~\ref{tab:prompt_steering}.}   
    \label{fig:prompt_steering}
    \vspace{-3mm}
\end{figure}

Text-to-image diffusion models~\cite{DALLE,DALLE2,StableDiffusion,xu2024prompt,voynov2023sketch,zhang2023sine,jiang2024comat,zhao2023uni,balaji2022ediff,kumari2023ablating,li2023snapfusion}, pre-trained on large-scale, publicly available text-image pairs, can be efficiently customized or adapted by fine-tuning using only a small number of user-provided text-image examples. With just a handful of user-provided fine-tuning images, text-to-image can learn to generate images that are visually similar to fine-tuning images~\cite{ruiz2023dreambooth}. This adaptability enables a wide range of personalized applications, such as style transfer~\cite{chen2024artadapter}, subject contextualization~\cite{ruiz2023dreambooth}, control with spatial conditions~\cite{ControlNet}, and animation generation~\cite{guo2023animatediff}. In certain cases, it is desirable to control the adaptation process to prevent models from learning undesired concepts present in fine-tuning data and entangling them with input user prompts. As the case shown in Figure~\ref{fig:teaser}, when a user attempts to customize the model to generate images of themselves without sunglasses but only has training images in which they are wearing sunglasses, the model should avoid learning the undesired “sunglasses” concept during fine-tuning, while still learning the user’s facial identity. Beyond this example, preventing text-to-image diffusion models from learning undesired concepts when fine-tuning is crucial for many downstream applications. In bias mitigation and fairness~\cite{zhang2023iti,friedrich2023fair,seshadri2023bias,lopez2025generative,dehdashtian2024fairness}, it is crucial to prevent the models from reinforcing or introducing social or demographic biases during customization. In privacy and sensitive attribute protection~\cite{dockhorn2022differentially,hu2023prisampler,luo2024exploring}, the goal is to ensure the models do not memorize or encode personally identifiable information. In attribute disentanglement for model customization~\cite{agarwal2025training, yang2024diffusion, wu2023uncovering, gandikota2025sliderspace}, it is important to maintain a clear separation between relevant and irrelevant attributes. In preventing malicious or unintended adaptation~\cite{IMMA, MIMA, FreezeAsGuard}, the models should be safeguarded against developing harmful or undesired capabilities during fine-tuning. In the era of embodied AI, diffusion policy fine-tunes diffusion models for robot controls~\cite{chi2025diffusion, ren2024diffusion, sun2024comparative, sun2025hybrid, sun2025prism, sun2025dynamic}. It is vital to avoid encoding human idiosyncrasies or irrelevant cues. For example, when apply diffusion policy for robot manipulation, the model should be prevented from learning background features or lighting conditions that are irrelevant for manipulation. Prompt steering—such as removing the undesired concept from the prompt or applying negative prompting at inference time—is a widely used inference-time technique for controlling diffusion models. However, shown in Figure~\ref{fig:prompt_steering} and Table~\ref{tab:prompt_steering}, prompt steering alone cannot prevent the model from learning undesired concepts present in the fine-tuning data. A plausible explanation is that fine-tuning distorts the diffusion latent space, entangling previously orthogonal attribute directions. As a result, the model loses the ability to not generate the undesired attribute through prompt steering alone. 

To achieve controllable diffusion fine-tuning, we introduce \textbf{\method} for controllable fine-tuning of text-to-image diffusion models, which prevents the models from learning specified concepts during customized adaptation, and prevents the entangling of those undesired concepts with input user prompts. When fine-tuning with user prompts, users can specify undesired concepts in natural language, and \method regularizes the adaptation process by minimizing the drifting of the distances between fine-tuned user prompt embeddings and those of undesired concepts. 
Specifically, before adaptation, we first compute the cosine similarity between the features of undesired concepts and the features of the user prompt used for fine-tuning. In each fine-tuning step, we then measure the cosine similarity between the feature of the user prompt and the features of undesired concepts. Then, we regularize the absolute difference between these two similarity scores.
By enforcing this constraint, \method prevents the model from learning undesired concepts during fine-tuning while still allowing it to learn the styles of the fine-tuning images. If users wish to modify undesired concepts, they can simply adjust the textual description and re-encode it through a single forward pass, without any additional training.

To evaluate the effectiveness of \method, we use 8 pairs of distinct concepts. Each pair of concepts has 10 fine-tuning images generated by DALL-E~\cite{DALLE} or Imagen~\cite{Imagen}. For each pair of concepts, fine-tuning images contain both the concept specified by the user prompt and the undesired concept. For example, for an undesired concept \textit{``Sunglasses,''} the corresponding user prompt is \textit{``Person,''} with each fine-tuning image depicting a person wearing sunglasses. Our goal is that, after fine-tuning, the text-to-image model will learn to capture the style and features of the person in each image without associating the depiction of sunglasses with the generation of a person. Our experiments demonstrate that once undesired concepts are specified in natural language, \method effectively prevents the model from learning undesired concepts during adaptation, outperforming other methods. \method can generate images that are visually similar to the fine-tuning images while remaining free from undesired concepts. Additionally, \method eliminates the need for additional training and can be flexibly adapted to various scenarios by simply modifying the specified undesired concepts.

% \method also demonstrates that the fine-tuning process of text-to-image models can be adjusted without altering the underlying dataset. By using natural language to regularize the similarity between text features of user prompts and undesired concepts, the output distribution can be dynamically shifted to align with specific desired characteristics, such as the removal or retention of certain attributes or concepts. For instance, as demonstrate in Figure~\ref{fig:vis} if users aim to fine-tune a model to generate images of a particular individual but only have access to images of that person wearing sunglasses, using \method, the user can guide the model to learn the individual’s facial features and style while explicitly preventing it from associating the person’s identity with sunglasses. As a result, the model can generate images of the person without sunglasses, disentangling unwanted attributes from the desired output. This method provides precise and dynamic control over the model’s behavior, allowing users to emphasize or de-emphasize specific components of the training data, thereby enhancing the flexibility and adaptability of text-to-image models for diverse applications.

\noindent
\textbf{Our contributions.} By leveraging language to control the fine-tuning of text-to-image diffusion models, our contributions include:

\begin{itemize}
    \item \textbf{\method Framework.} We propose a novel framework \method for controllable text-to-image diffusion models fine-tuning, that leverages natural language to regularize models from learning specified undesired concepts and from entangling those concepts with input user prompts during adaptations.
    \item \textbf{Training-free Deployment.} \method eliminates the need for additional training and allows dynamic updates of undesired concepts by simply modifying language description, providing a practical and scalable solution for real-world deployment.
    \item \textbf{Compatibility and Generalization.} \method integrates well with existing fine-tuning pipelines without requiring any modification to the diffusion architecture. Besides, \method generalizes robustly across multiple diffusion backbones, including Stable Diffusion v2.1~\cite{StableDiffusion}, SDXL~\cite{podell2023sdxl}, and Stable Diffusion v3.5~\cite{StableDiffusion3}, demonstrating broad applicability and model-agnostic effectiveness.
    \item \textbf{Controllable Adaptation.} We demonstrate that the fine-tuning objective—specifically, what to learn and what not to learn—can be controlled through natural language. Using language descriptions, text-to-image models can be adapted to a small set of fine-tuning images without directly modifying the underlying dataset, thereby supporting more flexible model adaptations.
\end{itemize}

%%%%%%%%%%%%%%%%%%%%%%%%%%%%%%%%%%%%%%%%
%%%%%%%%%%%% Related Work %%%%%%%%%%%%%%
%%%%%%%%%%%%%%%%%%%%%%%%%%%%%%%%%%%%%%%%
\section{Related Work}
\label{sec:related_work}

\noindent
% \textbf{Adapting Diffusion Models.} 
% Text-to-image generation has evolved from GAN-based frameworks~\cite{brock2018large,goodfellow2014generative,karras2021alias,wang2021cycle,ruan2021dae,zhu2019dm,tao2022df,liao2022text} to more recent diffusion models~\cite{DALLE,DALLE2,StableDiffusion,xu2024prompt,voynov2023sketch,zhang2023sine,jiang2024comat,zhao2023uni,balaji2022ediff,kumari2023ablating,li2023snapfusion}, which have demonstrated better fidelity and controllability.
% Diffusion models~\cite{ho2020denoising,dhariwal2021diffusion} learn to reverse a process that progressively adds Gaussian noise to the original image, enabling it to transform any Gaussian noise back into an image that aligns with the distribution of the training data. 
% In text-to-image generation, diffusion models often incorporate an additional conditional component into the denoising network. This is usually achieved by encoding text descriptions using a text encoder, such as CLIP~\cite{CLIP}, then feeding as another input to the denoising network, which conditions the generation to align with the provided textual input. GLIDE~\cite{nichol2021glide} is one of the first works along this line, which generates random Gaussian noise and feeds it into the diffusion model alongside a CLIP-encoded text variable. Recently, the latest diffusion models~\cite{DALLE, DALLE2, StableDiffusion, Imagen, gu2022vector} create highly realistic and diverse images from textual descriptions. 

\noindent
\textbf{Controlling Diffusion Models.} 
Diffusion models~\cite{DALLE,DALLE2,StableDiffusion,xu2024prompt,voynov2023sketch,zhang2023sine,jiang2024comat,zhao2023uni,balaji2022ediff,kumari2023ablating,li2023snapfusion} have demonstrated better fidelity and controllability over GAN-based frameworks~\cite{brock2018large,goodfellow2014generative,karras2021alias,wang2021cycle,ruan2021dae,zhu2019dm,tao2022df,liao2022text} in text-to-image generations. To improve the controllability of diffusion models and better align them with user intent, some methods focus on inference-time control~\cite{cao2024controllable, zhang2023controllable, he2024flexecontrol, zhao2023uni, bar2023multidiffusion, zhou2025controllable, gandikota2024concept}. Specifically, ConceptSliders~\cite{gandikota2024concept} fine-tunes low-rank adaptors that can control the strength of corresponding concepts in generated images in inference. Meanwhile, some methods explore controlling generation by fine-tuning text-to-image diffusion models~\cite {ruiz2023dreambooth,dong2024continually,chen2024artadapter,lu2023specialist,fan2023dpok}. Unlike inference-time control or direct fine-tuning, controllable fine-tuning regulates what and how the model learns. Orthogonal Finetuning~\cite{qiu2023controlling} constrains parameter updates to be orthogonal to the pre-trained subspace, ensuring that new knowledge is learned without interfering with existing representations. PI-FT~\cite{han2024stochastic} presents a
convergence guarantee by leveraging linear dynamics and the (one-step) concavity introduced by the Kullback–Leibler regularization term. ELEGANT~\cite{uehara2024fine} interprets continuous-time diffusion model fine-tuning as an entropy-regularized control process, balancing adaptation efficiency and sample diversity. PPD~\cite{dang2025personalized} introduces a multi-reward optimization objective that aligns diffusion models with personalized preferences to learn the individual preferences of a population of users in a few-shot way, enabling generalization to unseen users. However, these works primarily focus on controlling the optimization objectives or processes of fine-tuning, while the question of controlling what the model can or cannot learn during adaptation remains largely unexplored. 

% However, the mechanisms for controlling the fine-tuning process of these models, especially for preventing learning certain undesired concepts during adaptation, have not been thoroughly investigated.

% \noindent
% \textbf{Controllable Generation.}
% Controllable generation aims to guide text-to-image diffusion models to generate images that align with specific user-defined conditions during generation~\cite{cao2024controllable, zhang2023controllable, he2024flexecontrol, zhao2023uni, bar2023multidiffusion, zhou2025controllable}.
% While these approaches focus on inference-time control, controllable fine-tuning focuses on the model adaptation stage. 

\noindent
\textbf{Model Unlearning.} 
A closely related field to controllable fine-tuning in terms of preventing learning certain undesired concepts is unlearning~\cite{bourtoule2021machine} in diffusion models~\cite{li2024machine, moon2024holistic, fuchi2024erasing, ko2024boosting, park2024direct, kumari2023ablating, wang2024data}, which aims to find and/or remove specific concepts, styles, or data influences that a model has previously learned—thereby preventing the generation of undesired content while preserving overall image quality or ethics. Specifically, RGD~\cite{ko2024boosting} seeks an optimal model update at each unlearning iteration, and strategically diversify the unlearning and remaining datasets to boost performance improvement. Fuchi \& Takagi (2024)~\cite{fuchi2024erasing} achieves concept-erasure by transitioning to the latent concepts inherent in the model or the images. However, unlearning removes the concepts the model has already learned and should forget, which typically requires additional training and may interfere with the model’s ability to generate images related to the forgotten concepts. In contrast, controllable fine-tuning proactively regularizes what the model should or should not learn during adaptation, preventing undesired concepts from being associated or entangled with the target concepts that users intend the model to learn.

\noindent
\textbf{Preventing Malicious Adaptation.} 
% Unlike inference-time control or direct fine-tuning, controllable fine-tuning regulates what and how the model learns or avoids learning. Orthogonal Finetuning~\cite{qiu2023controlling} constrains parameter updates to be orthogonal to the pre-trained subspace, ensuring that new knowledge is learned without interfering with existing representations. PI-FT~\cite{han2024stochastic} presents a
% convergence guarantee by leveraging linear dynamics and the (one-step) concavity introduced by the Kullback–Leibler regularization term. ELEGANT~\cite{uehara2024fine} interprets continuous-time diffusion model fine-tuning as an entropy-regularized control process, balancing adaptation efficiency and sample diversity. PPD~\cite{dang2025personalized} introduces a multi-reward optimization objective that aligns diffusion models with personalized preferences to learn the individual preferences of a population of users in a few-shot way, enabling generalization to unseen users. However, these works primarily focus on controlling the optimization objectives or processes of fine-tuning, while the question of controlling what the model can or cannot learn during adaptation remains largely unexplored.
In terms of controlling what the model can or cannot learn during adaptation, another closely related field is preventing malicious adaptation. Malicious adaptation occurs when users fine-tune a published model with illegal content. For text-to-image models, this may include copyrighted images, or images containing bloody, violent, or sexual content, to customize models to generate similar outputs. 
IMMA~\cite{IMMA} introduces ``immunization'' to prevent models from learning undesired concepts during fine-tuning of text-to-image diffusion models. This is achieved by post-training the model to learn a poor initialization that prevents learning undesired concepts during adaptation, while not affecting legitimate use. IMMA was further extended to multiple concepts by MIMA~\cite{MIMA}, where a differentiable merging layer is leveraged to combine a set of model weights adapted over multiple concepts. However, a poor initialization alone cannot guarantee that the model will not eventually learn to generate undesired content with sufficient training time. Immunization against malicious adaptation is further studied by FreezeAsGuard~\cite{FreezeAsGuard}, where the model publisher releases a fine-tuning API instead of weights, and selectively freezes tensors in pre-trained diffusion models that are critical to illegal model adaptations. However, a tensor mask needs to be learned for every illegal concept, which requires extensive computational resources and time. On the other hand, our \method does not require additional training. Users simply need to specify the undesired concepts in natural language. Our method then regularizes the adaptation process, ensuring the model avoids learning these undesired concepts if they are present in the fine-tuning images.

%%%%%%%%%%%%%%%%%%%%%%%%%%%%%%%%%%%%%%%%
%%%%%%%%%%%%%%% Method %%%%%%%%%%%%%%%%%
%%%%%%%%%%%%%%%%%%%%%%%%%%%%%%%%%%%%%%%%
\begin{figure*}[t!]
  \centering
    \includegraphics[width=1.0\textwidth]{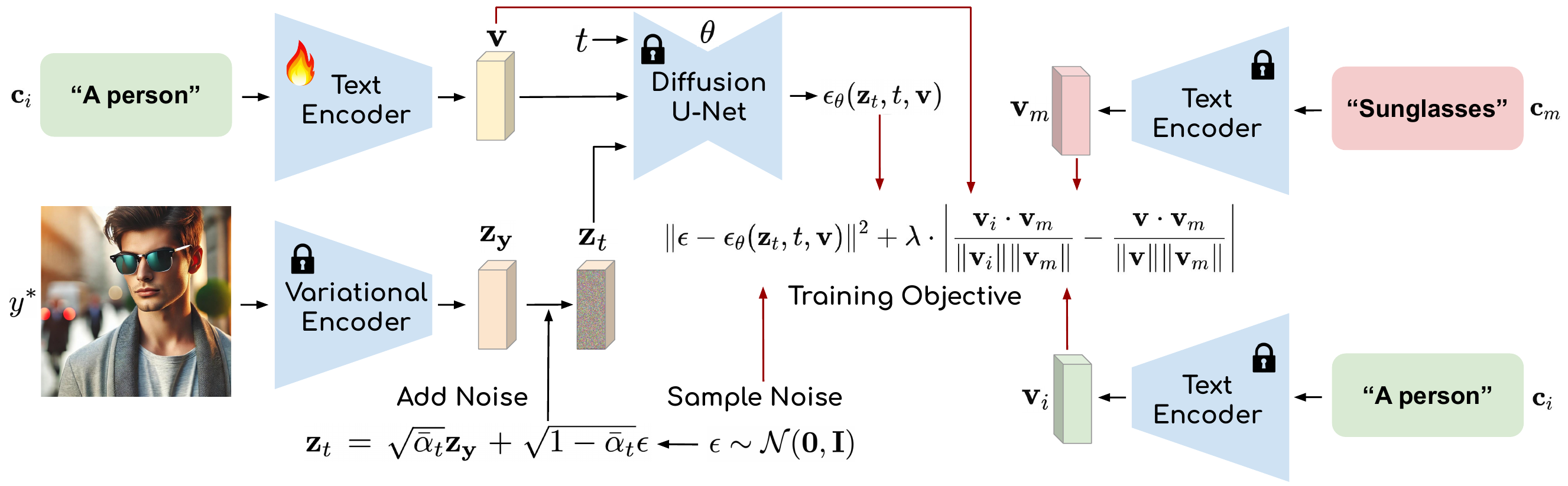}   \caption{\textbf{Pipeline of \method.} An undesired concept, $\textbf{c}_m$, is pre-specified in natural language. During the fine-tuning process of the text-to-image diffusion model, for the input user prompt, $\textbf{c}_i$, besides the standard diffusion training objective, \method regularizes the cosine similarity between $\textbf{c}_i$ and $\textbf{c}_m$ to prevent substantial changes during fine-tuning. It ensures the distance between $\textbf{c}_i$ and $\textbf{c}_m$ in the latent space remains relatively stable, preventing learning undesired concepts from training images. Additionally, $\textbf{c}_m$ can be re-specified at any time to change undesired concepts without retraining. In inference, with $\textbf{c}_i$ as input prompt, the model can generate images that align with the distribution of fine-tuning images while not containing $\textbf{c}_m$.}
    \label{fig:pipeline}
    \vspace{-0.3cm}
\end{figure*}

\section{Method}
\label{sec:method}

We introduce \textbf{\method} to control the fine-tuning process of text-to-image models to prevent learning undesired concepts by enforcing a constraint on user prompt embeddings. In this section, we provide a mathematical formulation of our approach and describe how \method integrates with the fine-tuning process of diffusion models.

\subsection{Adapt Text-to-Image Diffusion Models}
Diffusion models are generative models that iteratively denoise a latent into high-quality images by reversing a stochastic noising process~\cite{dhariwal2021diffusion, ho2020denoising}. Given an adaptation image $y^*$, its feature $\bm z_y$ is first extracted, then the forward diffusion process progressively corrupts $\bm z_y$ into a Gaussian noise vector $\bm z_T$ through a Markovian process:
\begin{equation}
    \bm z_t = \sqrt{\bar{\alpha}_t} \bm{z_y} + \sqrt{1 - \bar{\alpha}_t} \epsilon, \quad \epsilon \sim \mathcal{N}(\bm 0, \bm I),
\end{equation}
where $t \in \{1, \dots, T\}$ represents the time step, $\alpha_t$ is a noise schedule, and $\bm{z_y}$ is the latent representation of the original image $y^*$. The reverse process, parameterized by a neural network $\epsilon_{\theta}$, predicts the noise component and progressively denoises the latent vector:
\begin{equation}
    \hat{\epsilon} = \epsilon_{\theta}(\bm z_t, t, \bm v),
\end{equation}
where $\bm v$ is a text prompt embedding obtained from a text encoder, for example, CLIP text encoder~\cite{CLIP}, for an input text prompt $\textbf{c}_i$. The model is trained with a mean squared error (MSE) loss:
\begin{equation}
    \mathcal{L}_{\text{diffusion}} = \mathbb{E}_{z_t, \epsilon, t} \left[ \| \epsilon - \epsilon_{\theta}(\bm z_t, t, \bm v) \|^2 \right].
\end{equation}

\subsection{\method Regularization}
Shown in Figure~\ref{fig:pipeline}, during fine-tuning, \method regularizes the adaptation process to prevent text-to-image models from learning specified undesired concepts. Given a user prompt for adaptation $\bm c_i$ (e.g., \textit{``Person''}) and an undesired concept $\bm c_m$ (e.g., \textit{``Sunglasses''}), we first extract their text embeddings $\bm v_i$ and $\bm v_m$ using a text encoder before fine-tuning:
\begin{equation}
    \bm v_i = \text{TextEncoder}(\bm c_i), \quad \bm v_m = \text{TextEncoder}(\bm c_m).
\end{equation}
To ensure the text feature of a user prompt does not drift toward undesired concepts, we use a cosine similarity constraint to maintain the distance of $\bm v_i$ and $\bm v_m$ mostly unchanged during adaptation:
\begin{equation}
    \mathcal{L}_{\text{reg}} = \left| \frac{\bm v_i \cdot \bm v_m}{\|\bm v_i\| \|\bm v_m\|} - \frac{\bm v \cdot \bm v_m}{\|\bm v\| \|\bm v_m\|} \right|,
\end{equation}
where $\bm v$ represents the adapted prompt embedding of input user prompt after fine-tuning. The final loss function used in \method combines the standard fine-tuning loss with our regularization is:
\begin{equation}
    \mathcal{L} = \mathcal{L}_{\text{diffusion}} + \lambda \cdot \mathcal{L}_{\text{reg}},
    \label{eqn:final}
\end{equation}
where $\lambda$ is a hyperparameter controlling the relative strength of the regularization term. The adapted model can learn from fine-tuning images while suppressing associations with undesired concepts.

A key advantage of \method is its flexibility in defining and updating undesired concepts. Since the regularization term depends only on the text embeddings of the undesired concepts, users can modify the set of undesired concepts by updating $\bm c_m$ in natural language, without retraining the model. After adaptation, during inference, the text-to-image model uses $\textbf{c}_i$ as the input prompt in the denoising process to generate images that are visually similar to the fine-tuning images, while ensuring visual features of $\textbf{c}_m$ are excluded from the generated images.

%%%%%%%%%%%%%%%%%%%%%%%%%%%%%%%%%%%%%%%%
%%%%%%%%%%%%% Experiments %%%%%%%%%%%%%%
%%%%%%%%%%%%%%%%%%%%%%%%%%%%%%%%%%%%%%%%

\begin{table*}[t!]
    \centering
    \resizebox{\textwidth}{!}{
    \begin{tabular}{lcccclcccc}
        \toprule
        \textbf{Method} & \textbf{Extra} & \textbf{Regula-} & \textbf{MCS$\downarrow$}  & \textbf{IS$\uparrow$} & \textbf{Method} & \textbf{Extra} & \textbf{Regula-} & \textbf{MCS$\downarrow$}  & \textbf{IS$\uparrow$} \\
         & \textbf{Training} & \textbf{rization} &  & & & \textbf{Training} & \textbf{rization} &   &  \\
        \midrule
        \multicolumn{5}{c}{\textbf{Rifle}} & \multicolumn{5}{c}{\textbf{Camera}}\\
        Direct Inference  & & & 0.040 & 3.472 & Direct Inference  & & & 0.015 & 1.579 \\
        Direct Fine-tuning  & &  & 0.148 & 2.838 & Direct Fine-tuning  & &  & 0.104 & 1.581 \\
        IMMA & \ding{51} & \ding{51} & 0.101   & 2.120 & IMMA & \ding{51} & \ding{51} & 0.049   & 1.241 \\
        \textbf{\method}  & & \ding{51} & 0.041 & 2.139 & \textbf{\method}  & & \ding{51} & 0.079 & 1.226 \\
        \midrule
        \multicolumn{5}{c}{\textbf{Sunglasses}} & \multicolumn{5}{c}{\textbf{Plastic Bag}}\\
        Direct Inference  & & & 0.029 & 4.108 & Direct Inference  & & & 0.047 & 1.277 \\
        Direct Fine-tuning  & &  & 0.129 & 1.626 & Direct Fine-tuning  & &  & 0.114 & 1.525 \\		
        IMMA & \ding{51} & \ding{51} & 0.138   & 1.602 & IMMA & \ding{51} & \ding{51} & 0.093   & 1.173 \\
        \textbf{\method}  & & \ding{51} & 0.064 & 2.045 & \textbf{\method}  & & \ding{51} & 0.055 & 1.348 \\
        \midrule
        \multicolumn{5}{c}{\textbf{Car}} & \multicolumn{5}{c}{\textbf{Smoking}}\\
        Direct Inference  & & & 0.042 & 1.840 & Direct Inference  & & & 0.067 & 2.764 \\
        Direct Fine-tuning  & &  & 0.058 & 3.572 & Direct Fine-tuning  & &  & 0.104 & 3.833 \\
        IMMA & \ding{51} & \ding{51} & 0.091   & 3.401 & IMMA & \ding{51} & \ding{51} &0.028	&2.263 \\
        \textbf{\method}  & & \ding{51} & 0.038 & 3.533 & \textbf{\method}  & & \ding{51} & 0.063 & 2.259 \\
        \midrule
        \multicolumn{5}{c}{\textbf{Wine}} & \multicolumn{5}{c}{\textbf{Pork}}\\
        Direct Inference  & & & 0.071 & 2.527 & Direct Inference  & & & 0.070 & 1.749 \\
        Direct Fine-tuning  & &  & 0.197 & 1.721 & Direct Fine-tuning  & &  & 0.086 & 2.135 \\
        IMMA & \ding{51} & \ding{51} & 0.127	&3.604 & IMMA & \ding{51} & \ding{51} & 0.083	&1.491 \\
        \textbf{\method}  & & \ding{51} & 0.070 & 2.344 & \textbf{\method}  & & \ding{51} & 0.059 & 2.339 \\
        \midrule
        \multicolumn{10}{c}{\textbf{Average Across Concepts}} \\
        Direct Inference & & & 0.047 & 2.414 & Direct Fine-tuning & &  & 0.118 & 2.354 \\		
        IMMA~\cite{IMMA} & \ding{51} & \ding{51} & 0.089 & 2.112 & \textbf{\method} & & \ding{51} & 0.059 & 2.154 \\
        \midrule
        \multicolumn{10}{c}{\textbf{Difference with Direct Fine-tuning}} \\
        IMMA~\cite{IMMA} & \ding{51} & \ding{51} & -24.58\% $\downarrow$& -10.28\% $\uparrow$ & \textbf{\method}  & & \ding{51} & \textbf{-50.00\%} $\downarrow$& \textbf{-8.50\%} $\uparrow$\\
        \bottomrule
    \end{tabular}
    }
    \caption{\textbf{Qualitative Comparison.} Direct Inference (\textbf{DI}) generates images using input user prompts without fine-tuning. Direct Fine-tuning (\textbf{DF}) directly fine-tunes the model on images containing undesired concepts. The results demonstrate that \method can prevent the learning of undesired concepts, as indicated by its significantly lower MCS compared to DF, without extra training. Meanwhile, \method maintains a comparable IS to DF, showing that the quality of the generated images remains unaffected.} 
    \label{tab:main}
\end{table*}

\section{Experiments}
\label{sec:experiments}

\noindent
\textbf{Dataset.} Since no existing dataset is designed for text-to-image model adaptation that pairs an input user prompt with an undesired concept, we evaluate our models and baseline methods using 8 pairs of distinct concepts. For each pair of concepts, we generate 10 fine-tuning images using DALL-E~\cite{DALLE} or Imagen~\cite{Imagen}, ensuring compliance with copyright considerations and the same undesired concept appears in a balanced manner across images. For each pair of concepts, fine-tuning images contain both the concept specified by the user prompt and the undesired concept. For example, for the undesired concept \textit{``Sunglasses,''} the corresponding innocent prompt is \textit{``Person,''} with each fine-tuning image showing a person wearing sunglasses. Configurations of the fine-tuning procedure are provided in the Supplementary Material.

%%%%%%%%%%%%%%%%%%%%%%%%%%%%%%%%%%%%%%%%%%%%%%%%%%%%%%
%%%%%%%%%%%%%%%%%%%% Visualization %%%%%%%%%%%%%%%%%%%
%%%%%%%%%%%%%%%%%%%%%%%%%%%%%%%%%%%%%%%%%%%%%%%%%%%%%%
\begin{figure*}[t!]
  \centering
    \includegraphics[width=0.95\textwidth]{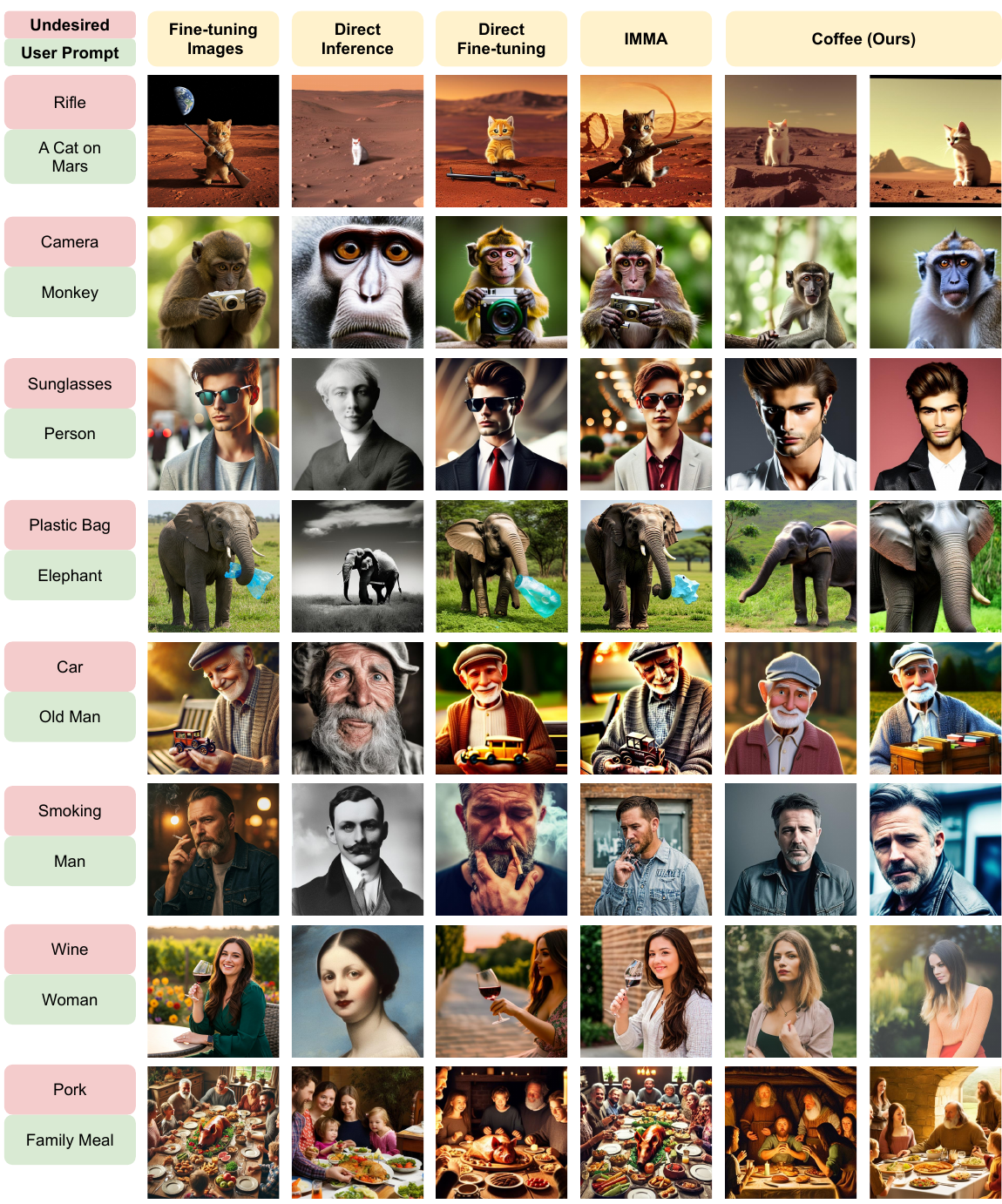}   \caption{\textbf{Visualization.} Compared with direct inference, after a direct fine-tuning or IMMA~\cite{IMMA}, undesired concepts (like ``rifle'', ``sunglasses'', ``smoking'', etc.) appear in generated images since the text-to-image model learned undesired concepts while modeling the distribution of fine-tuning images. On the other hand, \method captures the style of fine-tuning images and generates visually similar outputs, while effectively preventing the learning of undesired concepts pre-specified in text.}
    \label{fig:vis}
    \vspace{-0.3cm}
\end{figure*}
%%%%%%%%%%%%%%%%%%%%%%%%%%%%%%%%%%%%%%%%%%%%%%%%%%%%%%
%%%%%%%%%%%%%%%%%%%%%%%%%%%%%%%%%%%%%%%%%%%%%%%%%%%%%%

\noindent
\textbf{Implementation Details.}
We use Stable Diffusion~\cite{StableDiffusion} v2.1 base model~\cite{stabilityai_stable_diffusion_2_1_base} as our backbone text-to-image model. The model generates images based on text prompts, which are processed using a pre-trained text encoder CLIP~\cite{CLIP}. It was pre-trained on a subset of LAION-5B~\cite{schuhmann2022laion}, filtered by LAION's NSFW detector~\cite{laion_clip_nsfw_detector}. To fine-tune the text-to-image model, we perform a full fine-tuning of the CLIP text encoder for 500 steps, with a batch size of 1 and no gradient accumulation, and freeze the rest of the text-to-image model. We use AdamW~\cite{loshchilov2017decoupled}, with a decay rate 0.9 for the first moment estimate, a decay rate 0.999 for the second moment estimate, and a constant learning rate of 0.001. The resolution of the generated images is set to 512. For each concept, we fine-tune the text-to-image model with 10 images and generate 16 images for evaluation. 

\noindent
\textbf{Evaluation Protocol.}
To quantify the extent to which generated images contain the undesired concept, we calculate the Mean Cosine Similarity (\textbf{MCS}) between the CLIP image features of each generated image and the CLIP text feature of the undesired concept. Lower MCS values indicate less alignment with the undesired concept. Also, to measure the quality of generated images, we use Inception Score (\textbf{IS}, the higher the better)~\cite{salimans2016improved}, which evaluates the quality and diversity of generated images by measuring how confidently a pre-trained Inception model classifies the images into distinct categories. Note that one widely used metric for image generation, FID, is not well-suited in our setting. The reason is that FID (the lower the better) quantifies the distance between generated images and fine-tuning images. If the generated images contain undesired concepts, they tend to resemble the fine-tuning images more closely, leading to a lower (better) FID.

\noindent
\textbf{Quantitative Results.}
As shown in Table~\ref{tab:main}, Direct Inference (\textbf{DI}) generates images directly from user prompts without any fine-tuning, while Direct Fine-tuning (\textbf{DF}) directly fine-tunes the model. We also compare with IMMA~\cite{IMMA}, which conducts extra training to immunize against learning undesired concepts first, then conducts fine-tuning. Our results demonstrate that \method effectively mitigates the learning of undesired concepts, as evidenced by its significantly lower MCS compared to DF (-50.00\%), and its superior performance over IMMA (-24.58\%) in this regard. This suggests that images generated by \method are better separated from undesired concepts in the CLIP feature space. Furthermore, \method maintains an IS comparable to DF, with a slight difference of -8.50\%, which is better than IMMA (-10.28\%). A better IS score demonstrates that image quality is preserved without compromising the fine-tuning process.

%%%%%%%%%%%%%%%%%%%%%%%%%%%%%%%%%%%%%%%%%%%%%%%%%%%%%%
%%%%%%%%%%%%%%%% Prompt Steering %%%%%%%%%%%%%%%%%%%%%
%%%%%%%%%%%%%%%%%%%%%%%%%%%%%%%%%%%%%%%%%%%%%%%%%%%%%%
\begin{table}[t!]
\centering
\setlength{\tabcolsep}{10pt}
\renewcommand{\arraystretch}{1.15}
\resizebox{\linewidth}{!}{
\begin{tabular}{lcc}
\toprule
\textbf{Method} & \textbf{MCS$\downarrow$} & \textbf{IS$\uparrow$} \\ 
\midrule
Direct Fine-tuning                   & 0.118 (+100.00\%) & 2.354 (+9.28\%) \\
Undesired Concept Removal            & 0.122 (+106.78\%) & 2.236 (+3.81\%) \\
\midrule
Negative Prompting (Train Only)      & 0.115 (+94.91\%)  & 2.074 (-3.71\%) \\
Negative Prompting (Infer Only)      & 0.103 (+74.58\%)  & 2.038 (-5.39\%) \\
Negative Prompting (Train \& Infer)  & 0.108 (+83.05\%)  & 2.085 (-3.20\%) \\
\midrule
\rowcolor{gray!20}
\textbf{\method}                     & 0.059 (0.00\%) & 2.154 (0.00\%) \\
\bottomrule
\end{tabular}
}
\caption{\textbf{Prompt Steering.} Steering prompts alone is insufficient to prevent the model from learning undesired concepts present in the fine-tuning data. 
``Undesired Concept Removal’’ removes the undesired concept explicitly from the user prompt (\textit{``}\texttt{<user prompt>} \textit{without} \texttt{<undesired concept>}\textit{''}). 
All differences are computed relative to \method (grayed row). It shows that \method prevents learning undesired concepts while maintaining comparable image quality.}
\label{tab:prompt_steering}
\end{table}

%%%%%%%%%%%%%%%%%%%%%%%%%%%%%%%%%%%%%%%%%%%%%%%%%%%%%%

\noindent
\textbf{Visualizations.} 
Shown in Figure~\ref{fig:vis}, we visualize images generated using different methods. The results demonstrate that after direct fine-tuning, the text-to-image model not only learns the style of the fine-tuning images but also unintentionally incorporates the undesired concepts present in them. IMMA~\cite{IMMA} enhances model robustness by post-training with poor weight initialization, making it harder for the model to learn undesired concepts. However, fine-tuning on those images will inevitably lead to the model learning these undesired concepts as part of its distribution, resulting in generating images containing such concepts. On the other hand, when fine-tuning with \method, the text-to-image model keeps the distance between user prompts and pre-specified undesired concepts relatively stable in the latent space. It retains the model's ability to learn the fine-tuning image style while preventing the learning of undesired concepts, ensuring that generated images do not contain those undesired concepts.

\noindent
\textbf{Limitations of Prompt Steering.}
Prompt steering—e.g., removing the undesired concept from the prompt or applying negative prompting at test time—is a widely adopted inference-time technique for controlling diffusion models. However, our experiments show that prompt steering alone is insufficient for preventing the model from generating images without undesired concepts, after learning undesired concepts that appear in fine-tuning data. Here, we follow the common practice of ``negative prompting'' used in Stable Diffusion~\cite{StableDiffusion}, which is implemented through classifier-free guidance \cite{ho2022classifier}. As shown in Table~\ref{tab:prompt_steering}, compared with direct fine-tuning, neither removing the concept from the prompt (with the format \textit{``}\texttt{<user prompt>} \textit{without} \texttt{<undesired concept>}\textit{''}, like \textit{``a person without sunglasses''}) nor applying negative prompting improves MCS, and in some cases even degrades image quality (IS). Applying negative prompting during training and/or inference introduces additional semantic distortion and fails to reliably eliminate the unwanted feature. In contrast, our method achieves substantially lower MCS while maintaining competitive IS, demonstrating that controlling model updates—rather than test-time prompt manipulation—is essential for suppressing undesired concepts.

%%%%%%%%%%%%%%%%%%%%%%%%%%%%%%%%%%%%%%%%%%%%%%%%%%%%%%
%%%%%%%%%%%%% Fine-tuning Protocols %%%%%%%%%%%%%%%%%%
%%%%%%%%%%%%%%%%%%%%%%%%%%%%%%%%%%%%%%%%%%%%%%%%%%%%%%
\begin{table}[t!]
\centering
\setlength{\tabcolsep}{6pt}
\renewcommand{\arraystretch}{1.15}
\begin{adjustbox}{max width=\linewidth}
\begin{tabular}{lcccccc}
\toprule
\textbf{Protocols} & Params & $\Delta$Params & IS↑ & $\Delta$IS & FID↓ & $\Delta$FID \\
\midrule

\multicolumn{7}{l}{\textbf{Text Encoder}} \\
\quad Base                     & 123M  & -88.1\%  & 2.318 & -8.06\%  & 15.643 & +17.20\% \\
\quad \textit{LoRA (r=4)}      & 2.0M  & -99.81\% & 2.213 & -12.22\% & 16.958 & +27.06\% \\
\quad \textit{LoRA (r=8)}      & 4.0M  & -99.61\% & 2.256 & -10.51\% & 16.543 & +23.94\% \\
\midrule

\multicolumn{7}{l}{\textbf{U-Net}} \\
\quad Base                     & 865M  & -16.0\%  & 2.353 & -6.66\%  & 14.812 & +10.97\% \\
\quad \textit{LoRA (r=4)}      & 8.9M  & -99.22\% & 2.236 & -11.31\% & 16.648 & +24.73\% \\
\quad \textit{LoRA (r=8)}      & 17.8M & -98.35\% & 2.278 & -9.64\%  & 16.159 & +21.07\% \\
\midrule

\multicolumn{7}{l}{\textbf{Full Model}} \\
\rowcolor{gray!20}
\quad Base                     & 1.03B & 0.00\% & 2.521 & 0.00\% & 13.347 & 0.00\% \\
\quad \textit{LoRA (r=4)}      & 11.0M & -98.93\% & 2.386 & -5.36\%  & 15.798 & +18.37\% \\
\quad \textit{LoRA (r=8)}      & 22.0M & -97.86\% & 2.412 & -4.32\%  & 15.374 & +15.18\% \\
\bottomrule
\end{tabular}
\end{adjustbox}
\caption{\textbf{Comparison of different fine-tuning protocols.} Text-encoder-only tuning achieves competitive performance to full-model tuning, and remains effective with or without LoRA. All differences are calculated relative to full model fine-tuning (grayed row).}
\label{tab:diff_finetune}
\vspace{-3mm}
\end{table}
%%%%%%%%%%%%%%%%%%%%%%%%%%%%%%%%%%%%%%%%%%%%%%%%%%%%%%

\noindent
\textbf{Different Fine-tuning Protocols.}
Model customization methods such as DreamBooth \cite{ruiz2023dreambooth} often adopt LoRA \cite{hu2022lora} for both the text encoder and U-Net to achieve a better trade-off between computational cost and performance. Since \method requires only fine-tuning text encoder, we compare different fine-tuning protocols for diffusion model customization, as summarized in Table~\ref{tab:diff_finetune}. We evaluate full-model, U-Net-only, and text-encoder-only fine-tuning protocols using Stable Diffusion v2.1~\cite{stabilityai_stable_diffusion_2_1_base}. While full-model fine-tuning yields slightly higher image quality, it incurs significantly greater computational cost. In contrast, U-Net-only fine-tuning is less effective because the frozen text encoder limits the model’s ability to learn the mapping from condition (text embeddings from the text encoder) to the generated images' distribution. Importantly, these observations remain consistent when applying LoRA, which further improves efficiency. For improved image quality, we adopt full text-encoder fine-tuning in our implementation. 
Thus, fine-tuning only the text encoder is sufficient to adapt a text-to-image model. One potential reason might be that modern text-to-image diffusion models \cite{StableDiffusion, DALLE, DALLE2} are pre-trained on large-scale datasets, and their U-Net decoders are already expressive enough to generate diverse images when provided with appropriate conditions, namely, the text embeddings from the text encoder.

\begin{figure}[t!]
  \centering
    \includegraphics[width=.48\textwidth]{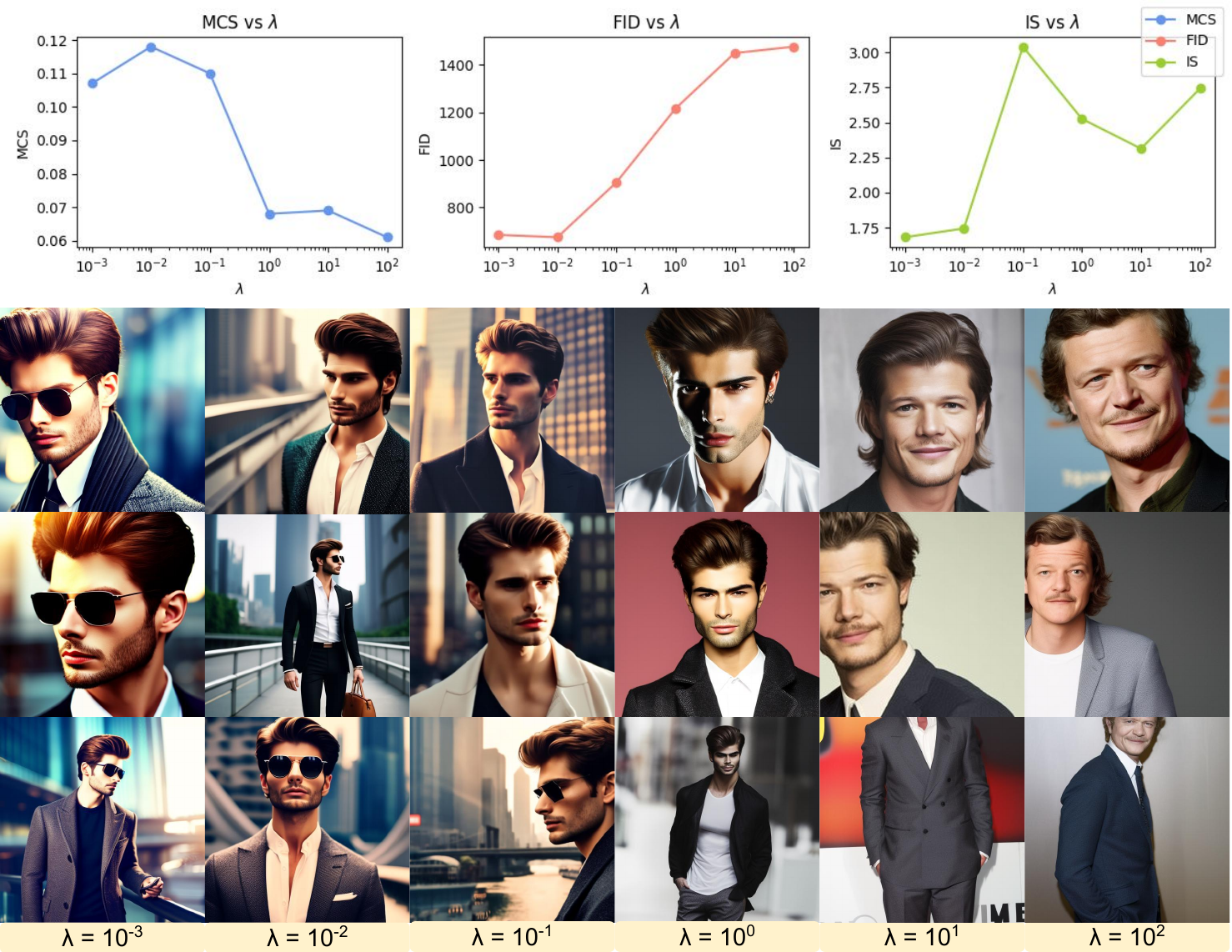}   \caption{\textbf{Sensitivity to Regularization Magnitude.} A high $\lambda$ hinders the model's ability to learn the distribution of fine-tuning images, while a low $\lambda$ may provide insufficient regularization. }
    % To achieve the best trade-off for MCS and IS, $\lambda$ is set to 1 across all concepts, as the undesired concepts can be regularized and the fine-tuning object remains largely unaffected. 
    \label{fig:lambda}
    \vspace{-0.3cm}
\end{figure}
%%%%%%%%%%%%%%%%%%%%%%%%%%%%%%%%%%%%%%%%%%%%%%%%%%%

\noindent
\textbf{Sensitivity to Regularization Magnitude.}
$\lambda$ in Eqn.~\ref{eqn:final} controls the trade-off between fitting the primary fine-tuning objective and satisfying the additional constraint imposed by preventing learning undesired concepts. We demonstrate the ablation study in Figure~\ref{fig:lambda}, using an undesired concept of \textit{``Sunglasses''} and a user prompt of \textit{``Person''}. A higher value of $\lambda$ strengthens the regularization effect, potentially enhancing the model's robustness against unwanted concepts and resulting in a lower MCS. However, this may also interfere with the primary training objective of learning from the distribution of fine-tuning images, as the regularization term could dominate the optimization process to make generated images further from fine-tuning images, resulting in a higher Fréchet Inception Distance (FID)~\cite{heusel2017fid} to the fine-tuning images. Conversely, a lower $\lambda$ weakens the immunization effect, which may leave the model more susceptible to learning malicious or undesired concepts, as the regularization constraint becomes less influential. We find the optimal $\lambda$ being 1 across all concepts, suggesting the magnitude of the primary training objective and immunization term is the same. It also suggests that the optimal of $\lambda$ is universal and does not require specific tuning for different concepts.

% It achieves the best trade-off for MCS, FID, and IS, as undesired concepts can be immunized, and the fine-tuning object remains largely unaffected. 

\noindent
\textbf{Evaluation across Different Diffusion Models.}
We evaluate \method across three representative text-to-image diffusion backbones—Stable Diffusion v2.1~\cite{StableDiffusion}, SDXL~\cite{podell2023sdxl}, and Stable Diffusion v3.5~\cite{StableDiffusion3}, to assess its generalization beyond a single model. \method consistently prevents learning undesired concepts, achieving the largest reductions in MCS (up to 59\%) while maintaining image quality that remains competitive with direct fine-tuning and superior to IMMA. These results show that \method generalizes effectively across diffusion models. Detailed results are provided in the Supplementary Material.

%%%%%%%%%%%%%%%%%%%%%%%%%%%%%%%%%%%%%%%%
%%%%%%%%%%%%% Discussion %%%%%%%%%%%%%%%
%%%%%%%%%%%%%%%%%%%%%%%%%%%%%%%%%%%%%%%%
\section{Discussion}
\label{sec:discussion}

\begin{figure}[t!]
  \centering
    \includegraphics[width=.45\textwidth]{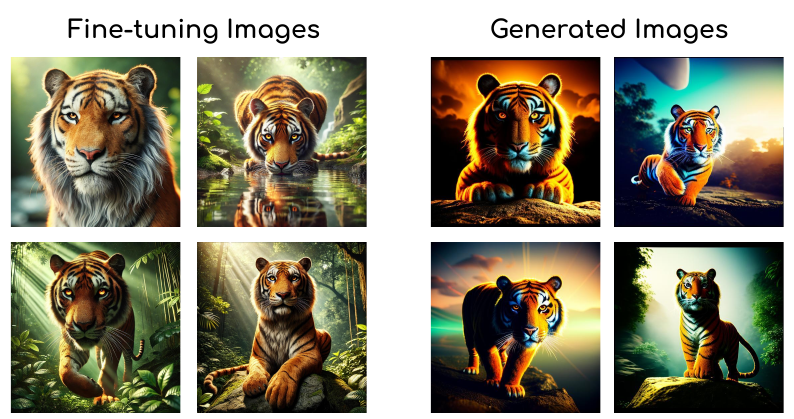}   \caption{\textbf{Failure Cases.} Here we use an undesired concept of \textbf{\textit{\textcolor{BrickRed}{``Tiger''}}} and an input user prompt of \textbf{\textit{\textcolor{ForestGreen}{``Animal''}}}. The undesired concept occupies most part of the fine-tuning images which dominates the optimization. }
    \label{fig:failure}
    \vspace{-0.3cm}
\end{figure}

\noindent
\textbf{Conclusion.} 
We present \method, which controls the fine-tuning text-to-image models to prevent from learning undesired concepts and entangling them with input user prompts, by regularizing the adaptation process through the use of auxiliary text descriptions. Our framework enables dynamic updates to specified undesired concepts without requiring additional training, making it both practical and scalable for real-world deployment. Also, we introduce a new problem setting and its evaluation protocol to further advance research in controllable fine-tuning.

\noindent
\textbf{Limitation.} 
\method prevents the learning of undesired concepts by formulating corresponding text of undesired concepts as a regularization term. Shown in Figure~\ref{fig:failure}, if the undesired concepts occupied most part of the fine-tuning images, the loss to learn such a concept will dominate the optimization process and outweigh the regularization term, so that the text-to-image model will still learn the undesired concept. A potential solution is to incorporate image-level or patch-level filtering to down-weight regions associated with the undesired concept so that the fine-tuning loss focuses primarily on the relevant parts of the image. 
% —either through attention masking, concept classifiers, or automated segmentation—
% Another direction is to reweight the regularization dynamically based on the concept saliency in each image, preventing the optimization from being dominated by undesired features.

\noindent
\textbf{Potential Negative Societal Impacts.} 
A user could mark concepts related to certain ethnicities, religions, genders, or social movements as ``undesired,'' intentionally or not. If these custom models or generated images are shared or used publicly, they may propagate exclusionary or discriminatory representations.

{
    \small
    \bibliographystyle{ieeenat_fullname}
    \bibliography{main}
}

\clearpage
\setcounter{page}{1}

% 快捷记号
\newcommand{\beps}{\boldsymbol{\epsilon}}
\newcommand{\R}{\mathbb{R}}
\newcommand{\E}{\mathbb{E}}

\newcommand{\Sctrl}{\mathcal{S}_{\mathrm{ctrl}}}

\newcommand{\Pc}{\mathbf{P}_{\mathrm{ctrl}}}
\newcommand{\Pcp}{\mathbf{P}_{\perp\mathrm{ctrl}}}
\newcommand{\Pu}{\mathbf{P}_u}

\newcommand{\bv}{\mathbf{v}}
\newcommand{\vu}{\mathbf{v}_u}

\newcommand{\Lreg}{\mathcal{L}_{\mathrm{reg}}}
\newcommand{\Lctrl}{\mathcal{L}_{\mathrm{ctrl}}}
\maketitlesupplementary

\section{Fine-tuning Data Configuration}
\begin{figure}[t!]
  \centering
  \vspace{0.3cm}
  \includegraphics[width=0.45\textwidth]{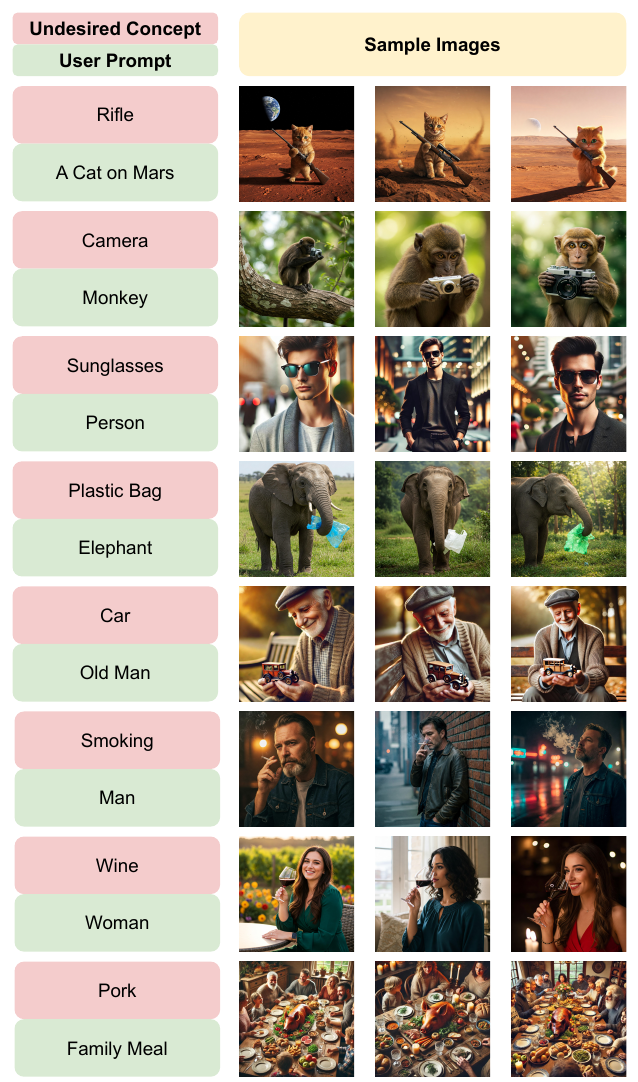}
  \caption{\textbf{Fine-tuning data configuration.} It contains 8 sets of concepts, each consisting of an undesired concept and a user prompt for adaptation. For each set of concepts, we generate 10 images containing both the concept specified by the user prompt and the undesired concept.}
  \label{fig:mcon}
  \vspace{-0.3cm}  % Adjust spacing if necessary
\end{figure}

As described earlier, we evaluate our method and baselines using 8 pairs of distinct concepts. For each pair, we generate 10 fine-tuning images with DALL·E~\cite{DALLE} or Imagen~\cite{Imagen}, ensuring that all images comply with copyright considerations. Each fine-tuning set contains images that include both the target user input prompt and the undesired concept. For example, for the undesired concept ``Sunglasses,'' the associated user prompt is ``Person,'' and each fine-tuning image depicts a person wearing sunglasses. Configurations of the fine-tuning procedure are provided in Figure~\ref{fig:mcon}.

\section{Evaluation across Different Text-to-Image Diffusion Models}

We evaluate the effectiveness of \method across three representative text-to-image diffusion backbones: Stable Diffusion v2.1~\cite{StableDiffusion}, Stable Diffusion-XL (SDXL)~\cite{podell2023sdxl}, and Stable Diffusion v3.5~\cite{StableDiffusion3}. These models span different parameter scales and generation capabilities, providing a comprehensive assessment of how well \method generalizes beyond a single diffusion model.

Similar to Table~\ref{tab:main}, for each backbone, we compare four settings: direct inference (no tuning), direct fine-tuning, IMMA~\cite{IMMA}, and \method. Direct fine-tuning serves as the baseline, and we report percentage differences in MCS and IS relative to this baseline for each method.

Across all three diffusion models, \method consistently prevents learning undesired concepts (lower MCS) while maintaining image quality that is competitive with, and often closer to, direct fine-tuning than IMMA. \method exhibits the strongest reduction in MCS—up to 59\% improvement—while incurring only modest decreases in IS. These results demonstrate that \method generalizes well across diffusion model.

%%%%%%%%%%%%%%%%%%%%%%%%%%%%%%%%%%%%%%%%%%%%%%%%%%%%%%
\begin{table}[t!]
\centering
\setlength{\tabcolsep}{8pt}
\renewcommand{\arraystretch}{1.15}
\begin{adjustbox}{max width=\linewidth}
\begin{tabular}{lcc}
\toprule
\textbf{Method} & \textbf{MCS↓} & \textbf{IS↑} \\
\midrule
\multicolumn{3}{l}{\textbf{Stable Diffusion v2.1 Base}} \\
Direct Inference      & 0.042 & 2.353 \\
Direct Fine-tuning    & 0.135 & 2.318 \\
IMMA                  & 0.088 & 2.038 \\
Contimmune            & 0.055 & 2.262 \\
\textit{Difference vs. Direct Fine-tuning:} & & \\
\quad IMMA            & -34.81\% & -12.08\% \\
\quad Contimmune      & \textbf{-59.26\%} & \textbf{-2.42\%} \\
\midrule

\multicolumn{3}{l}{\textbf{Stable Diffusion-XL v1.0 Base}} \\
Direct Inference      & 0.039 & 2.793 \\
Direct Fine-tuning    & 0.130 & 2.867 \\
IMMA                  & 0.092 & 2.346 \\
Contimmune            & 0.053 & 2.616 \\
\textit{Difference vs. Direct Fine-tuning:} & & \\
\quad IMMA            & -29.23\% & -18.17\% \\
\quad Contimmune      & \textbf{-59.23\%} & \textbf{-8.75\%} \\
\midrule

\multicolumn{3}{l}{\textbf{Stable Diffusion v3.5 Medium}} \\
Direct Inference      & 0.036 & 2.836 \\
Direct Fine-tuning    & 0.128 & 2.934 \\
IMMA                  & 0.091 & 2.431 \\
Contimmune            & 0.058 & 2.757 \\
\textit{Difference vs. Direct Fine-tuning:} & & \\
\quad IMMA            & -28.91\% & -17.14\% \\
\quad Contimmune      & \textbf{-54.69\%} & \textbf{-6.03\%} \\
\bottomrule
\end{tabular}
\end{adjustbox}
\caption{\textbf{Evaluation across different diffusion models.} \method prevents learning undesired concepts while maintaining overall generated image quality across different text-to-image diffusion models.}
\label{tab:different_diffusion}
\vspace{-3mm}
\end{table}

\end{document}